\documentclass[runningheads]{llncs}

\usepackage{eccv}

\usepackage{eccvabbrv}

\usepackage{graphicx}
\usepackage{booktabs}
\usepackage{multirow}
\usepackage{algorithm}
\usepackage{algorithmic}

\usepackage{pifont}
\usepackage{bbding}
\usepackage{multirow}
\usepackage[accsupp]{axessibility}  
\renewcommand{\mathbf}{\boldsymbol}
\def\x{\mathbf{x}}

\def\ll{\mathbf{l}}
\def\tt{\mathbf{t}}
\def\cc{\mathbf{c}}
\def\p{\mathbf{p}}
\def\C{\mathbf{C}}

\def\vv{\mathbf{v}}
\def\V{\mathbf{V}}

\usepackage[dvipsnames]{xcolor}
\usepackage{color, colortbl}

\definecolor{LightCyan}{rgb}{0.88,1,1}
\definecolor{HighLight}{rgb}{0.96,0.92,0.96}

\usepackage{hyperref}

\usepackage{orcidlink}

\begin{document}


\title{LAPT: Label-driven Automated Prompt Tuning for OOD Detection with Vision-Language Models} 

\titlerunning{LAPT for OOD Detection with Vision-Language Models}

\author{Yabin Zhang\inst{1,2}\and
Wenjie Zhu\inst{1} \and
Chenhang He\inst{1} \and
Lei Zhang\inst{1,2}\thanks{Corresponding author.}}

\authorrunning{Y. Zhang et al.}

\institute{The Hong Kong Polytechnic University \and
OPPO Research Institute
\\
{\{csybzhang,cslzhang\}@comp.polyu.edu.hk} 
}

\maketitle

\begin{abstract}

Out-of-distribution (OOD) detection is crucial for model reliability, as it identifies samples from unknown classes and reduces errors due to unexpected inputs. 
Vision-Language Models (VLMs) such as CLIP are emerging as powerful tools for OOD detection by integrating multi-modal information.
However, the practical application of such systems is challenged by manual prompt engineering, which demands domain expertise and is sensitive to linguistic nuances.
In this paper, we introduce \textbf{L}abel-driven \textbf{A}utomated \textbf{P}rompt \textbf{T}uning (LAPT), a novel approach to OOD detection that reduces the need for manual prompt engineering.
We develop distribution-aware prompts with in-distribution (ID) class names and negative labels mined automatically. Training samples linked to these class labels are collected autonomously via image synthesis and retrieval methods, allowing for prompt learning without manual effort.
We utilize a simple cross-entropy loss for prompt optimization, with cross-modal and cross-distribution mixing strategies to reduce image noise and explore the intermediate space between distributions, respectively.
The LAPT framework operates autonomously, requiring only ID class names as input and eliminating the need for manual intervention.
With extensive experiments, LAPT consistently outperforms manually crafted prompts, setting a new standard for OOD detection.
Moreover, LAPT not only enhances the distinction between ID and OOD samples, but also improves the ID classification accuracy and strengthens the generalization robustness to covariate shifts, resulting in outstanding performance in challenging full-spectrum OOD detection tasks. Codes are available at \url{https://github.com/YBZh/LAPT}.

\end{abstract}

\section{Introduction}
\label{sec:intro}

In real-world applications, artificial intelligence (AI) systems often encounter data of unknown classes, known as out-of-distribution (OOD) data. When tackling such OOD data, AI models may erroneously remain overconfident in their predictions \cite{scheirer2012toward,nguyen2015deep}, resulting in critical errors and posing security risks. 
Consequently, it is crucial to recognize OOD data for maintaining the reliability and safety of AI systems in open-world scenarios.

\begin{figure}[tb]
    \centering
\includegraphics[width=0.78\columnwidth]{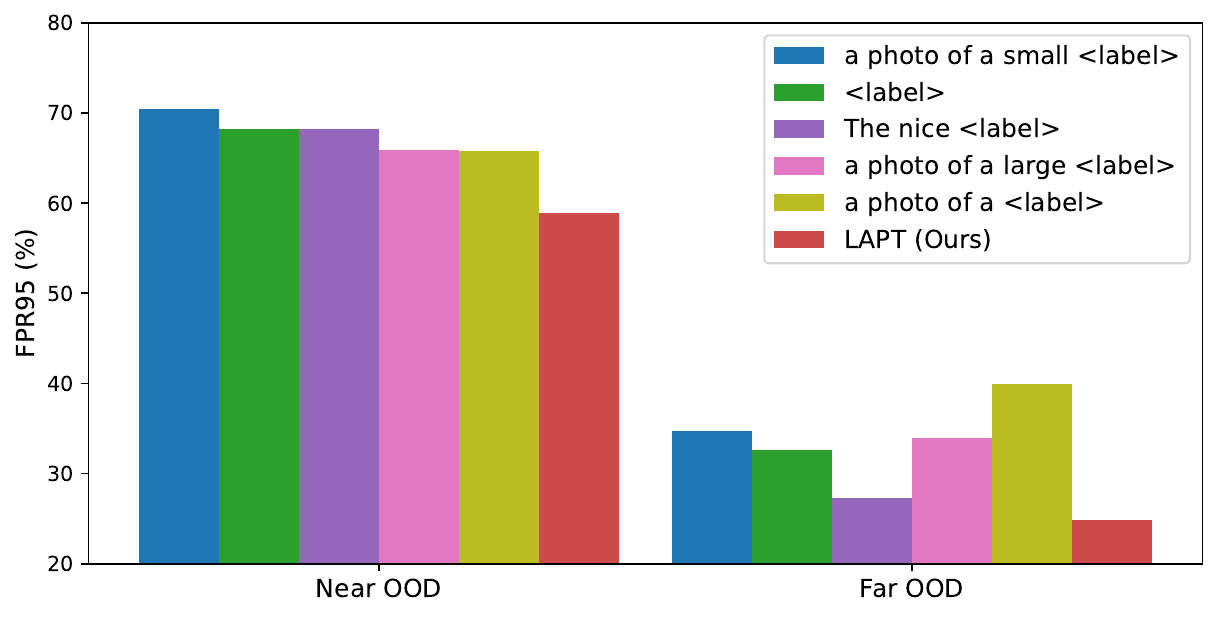}
    \caption{VLMs-based OOD detection systems are sensitive to linguistic nuances of text prompts, where results are reported on the OpenOOD dataset with a ViTB/16 image encoder. 
    A lower FPR95 metric denotes better OOD detection performance.}
    \label{fig:subtle_text_change}
\end{figure}

Traditional OOD detection methods are primarily focused on image information \cite{hendrycks2016baseline,lee2018simple,liang2017enhancing}, neglecting the rich textual knowledge carried by class names. With the rise of vision-language models (VLMs) \cite{radford2021learning,jia2021scaling}, there has been a growing interest in leveraging textual information to facilitate visual tasks \cite{zhou2022learning,zhang2021tip,zhang2024dual}, including OOD detection \cite{ming2022delving,ming2024does,jiang2024negative}. For example, Ming \emph{et al.} \cite{ming2022delving} compared test images with linguistic descriptions of in-distribution (ID) classes in feature space, leveraging the maximum probability of a scaled softmax function to identify OOD images. While this method achieves notable generalization across various datasets, it engages only a restricted portion of the textual space, thereby not fully exploiting the textual interpretative capabilities of VLMs.
Jiang \emph{et al.} \cite{jiang2024negative} extended it by exploiting larger textual spaces, extracting negative labels from vast text corpora to enhance OOD detection performance significantly. However, the deployment of this system is hindered by the demanding task of prompt engineering, which requires domain-specific knowledge and is sensitive to fine-grained linguistic variations, as illustrated in \cref{fig:subtle_text_change}.

To this end, we propose \textbf{L}abel-driven \textbf{A}utomated \textbf{P}rompt \textbf{T}uning (LAPT), a novel approach to OOD detection that substantially reduces the need of manual prompt engineering.
Specifically, given ID class names and  automatically extracted negative labels as in \cite{jiang2024negative}, 
we append distribution-aware prompt tokens to these class labels. 
These learnable tokens are optimized with minimal human intervention by aggregating training images associated with class labels using pre-trained text-to-image generation models \cite{rombach2022high,podell2023sdxl} or retrieving real images from web-scale dataset \cite{schuhmann2021laion,schuhmann2022laion}.
We employ a simple yet effective cross-entropy loss for prompt optimization, enhanced by cross-modal and cross-distribution data mixing techniques.
Cross-modal mixing merges image and text features from the same class, mitigating image noise through multi-modal integration.
Cross-distribution mixing stochastically blends ID and negative features along with their respective labels, exploring the intermediate space between ID and negative domains. During testing, we adopt the OOD score introduced in \cite{jiang2024negative} to utilize both ID and negative labels.


Extensive experiments demonstrate that our method outperforms the manually engineered text prompts by a large margin and achieves new state-of-the-art on various OOD detection benchmarks. Especially, on the challenging near-OOD detection task using the ImageNet-1k as the ID dataset, our approach exceeds the best hand-crafted prompts by 10.76\% in FPR95 and 5.71\% in AUROC, without any manual label annotations or prompt crafting. 
Moreover, LAPT not only enhances the distinction between ID and OOD samples, but also improves the ID classification accuracy and strengthens the generalization robustness to covariate shifts.
These improvements collectively contribute to the superior performance of LAPT in challenging full-spectrum OOD detection tasks \cite{yang2023full,yang2021generalized}.
We summarize our contributions as follows:
\begin{itemize}
    \item To circumvent the labor-intensive task of manual prompt engineering for OOD detection with VLMs, we introduce the LAPT methodology, which autonomously learns effective prompts using only ID class names.
    \item Utilizing the ID class names and automatically mined negative labels, we devise distribution-aware prompts and autonomously gather the corresponding training images for each class via image synthesis and retrieval.    
    \item With the collected training data, we employ a simple yet effective cross-entropy objective for prompt optimization, enhanced by cross-modal data mixing to reduce image noise and cross-distribution data mixing to explore the intermediate space between ID and negative regions.    
    \item Extensive experiments demonstrate that our method significantly outperforms the hand-crafted text prompts and achieves new state-of-the-art results across multiple benchmarks. Moreover, LAPT enhances not only the distinction between ID and OOD samples, but also the ID classification accuracy and the generalization robustness to covariate shifts, achieving outstanding results in full-spectrum OOD detection tasks.
\end{itemize}

\section{Related Work}
\textbf{Traditional OOD detection} aims to address the severe issue of deep learning models in producing overconfident predictions on OOD data \cite{nguyen2015deep,yang2021generalized}, which poses significant challenges to the deployment of deep models in real-world applications. To tackle this problem, researchers have developed various OOD detection methods, including score-based \cite{hendrycks2016baseline,lee2018simple,liang2017enhancing,liu2020energy,wang2021energy,huang2021mos,wang2022vim,wei2022mitigating,sun2021react}, distance-based \cite{tack2020csi,tao2023non,sun2022out,du2022siren,ming2022exploit,sehwag2021ssd} and generative-based \cite{ryu2018out,kong2021opengan} approaches.
Score-based methods, which arguably garner the most attentions, distinguish ID from OOD samples using carefully designed scoring functions, such as confidence-based \cite{hendrycks2016baseline,liang2017enhancing,sun2021react,wang2022vim,wei2022mitigating}, discriminator-based \cite{kong2021opengan}, energy-based \cite{liu2020energy,wang2021energy}, and gradient-based \cite{huang2021importance} scores. In contrast, distance-based methods detect OOD samples by measuring the distance in feature space between the test data and their nearest ID samples \cite{tack2020csi} or ID prototypes \cite{tao2023non} using metrics like KNN \cite{sun2022out,du2022siren,ming2022exploit} or Mahalanobis distance \cite{lee2018simple,sehwag2021ssd}.
While these methods have achieved certain success, they overlook textual information and require training or pre-training on manually labeled ID training data. Moreover, some approaches require manually collecting OOD data for additional training \cite{fort2021exploring,hendrycks2018deep,zhang2021fine}, hindering their real-world applicability.
Our method differs from these approaches by effectively leveraging textual information. 
Utilizing only the ID class names, we mine negative labels, gather training data, and learn effective prompts in an automated fashion, achieving state-of-the-art OOD detection results with minimal human effort.



\textbf{OOD detection with VLMs} is gaining growing attention as the importance of textual information for OOD image detection becomes increasingly recognized \cite{ming2022delving,ming2024does,jiang2024negative,esmaeilpour2022zero,miyai2024locoop,wang2023clipn,nie2023out}.
MCM \cite{ming2022delving} leverages ID class names for effective zero-shot OOD detection, which is further enhanced by NegLabel \cite{jiang2024negative} through the additional use of mined OOD class names. However, as shown in \cref{fig:subtle_text_change}, NegLabel's performance is sensitive to text prompts, limiting its practical application.
ZOC \cite{esmaeilpour2022zero} employs VLMs to detect OOD samples by learning a captioner that generates candidate OOD labels. Yet, this captioner struggles to generate effective OOD labels for large-scale ID datasets with numerous ID classes.
LoCoOp \cite{miyai2024locoop} learns ID prompts using few-shot ID samples, regularizing these ID prompts  by mined OOD features from image backgrounds.
CLIPN \cite{wang2023clipn} and LSN \cite{nie2023out} learn text prompts for OOD classes. Our approach differs significantly from them in two aspects. First, we initialize negative prompts with negative labels instead of initializing negative prompts with 'no' and ID labels. Considering CLIP's difficulty in understanding the meaning of 'no' \cite{nie2023out}, using OOD labels to initialize negative prompts provides an effective starting point to accelerate convergence. Second, CLIPN and LSN respectively require large-scale multi-modal data and manually collected ID samples, leading to either high training costs or the need of manual annotation. In contrast, our method automatically gathers a small amount of positive/negative training data with ID labels, requiring minimal manual effort yet achieving improved performance. 

\section{Methods}

\subsection{Problem Setup}
Let $\mathcal{X}$ and $\mathcal{Y}=\{ y_1, \dots, y_C \}$ be the image space and ID label space, where $\mathcal{Y}$  is a set of class names,  \eg, $\mathcal{Y}=\{ cat, dog, \dots, bird \}$ and $C$ is the number of ID classes. Given the ID random variable $\x^{in} \in \mathcal{X}$ and the OOD random variable $\x^{ood} \in \mathcal{X}$, we use $\mathcal{P}_{\x^{in}}$ and $\mathcal{P}_{\x^{ood}}$ to denote the ID marginal distribution and OOD marginal distribution, respectively.

\textbf{OOD detection.} In traditional classification tasks, we assume that the test image $\x$ is sampled from the ID distribution and belongs to a specific ID class, \ie, $\x \in\mathcal{P}_{\x^{in}}$ and $y \in \mathcal{Y}$, where $y$ is the label of $\x$.
However, in real-world applications, AI systems often encounter samples from unknown classes, \ie, $\x \in\mathcal{P}_{\x^{ood}}$ and $y \notin \mathcal{Y}$. AI models may misclassify these data into known ID classes with high confidence \cite{scheirer2012toward,nguyen2015deep}, posing critical errors and security risks. 
To tackle this problem, OOD detection is introduced to classify ID samples into correct ID classes and reject OOD samples simultaneously. The ID class classification is achieved with a $C$-way classifier following traditional classification \cite{krizhevsky2012imagenet,he2016deep}, while the OOD detection is typically achieved with a score function $S$ \cite{lee2018simple,liang2017enhancing,liu2020energy}:
\begin{equation}
    G_{\gamma}(\x) = \textrm{ID, } \textrm{if } S(\x) \geq \gamma; \quad \textrm{otherwise, } G_{\gamma}(\x) = \textrm{OOD},
\end{equation}
where $G_{\gamma}$ is the OOD detector with the threshold $\gamma$. In other words, the test data $\x$ is detected as ID samples if and only if $S(\x) \geq \gamma$.

\subsection{Reviews on MCM and NegLabel}

\textbf{CLIP-like models.} Given the ID test image $\x$ within the label space $\mathcal{Y}$, we extract the image representation $\vv = f_{img}(\x) \in \mathcal{R}^D$ and the textual representation $\C = f_{txt}(\rho(\mathcal{Y})) \in \mathcal{R}^{C\times D}$ with the pre-trained CLIP-like encoders, where $D$ denotes the feature dimension. $f_{img}(\cdot)$ and $f_{txt}(\cdot)$ denote the image and text encoders, respectively.
$\rho(\cdot)$ denotes the text prompt function, which is typically manually defined as `a photo of a <label>.', where the `label' token is replaced by the specific class name, \eg, `cat' or `dog'.
Both $\vv$ and $\C$ are $L_2$ normalized along the $D$ dimension. Then the zero-shot classification probability can be achieved using $\C$ as the classifier:
\begin{equation}
    \p = \mathrm{Softmax}(\vv\C^T) \in \mathcal{R}^{C},
\end{equation}
where the fixed scaling parameter is omitted for simplicity.

\textbf{MCM.} 
Besides the remarkable classification capabilities, MCM \cite{ming2022delving} reveals that pre-trained CLIP models also exhibit strong zero-shot OOD detection abilities. Specifically, MCM treats textual representations $\C$ as ID prototypes and assesses OOD uncertainty based on the scaled cosine distance between the visual features and the nearest ID prototype. Furthermore, it has been validated that applying a softmax function as a post hoc mechanism enlarges the distinction between ID and OOD samples, leading to the following score function:
\begin{equation} \label{equ:mcm_score}
    S_{mcm}(\vv) = \max_i \frac{e^{\cos(\vv, \cc_i)/\tau}}{\sum_{i=1}^C {e^{\cos(\vv, \cc_i)/\tau}}},
\end{equation}
where $\mathrm{cos}(\cdot,\cdot)$ measures the cosine similarity, $\cc_i$ is the $i$-th entry of $\C$, and $\tau > 0$ is the scaling temperature. 

\textbf{NegLabel.}
While it is validated in \cite{ming2022delving} that leveraging the textual information of ID classes enhances the OOD detection capability, the potential contribution of negative labels is overlooked. 
To fill this gap, NegLabel \cite{jiang2024negative} explores negative class names $\mathcal{Y}^- =\{ y_{C+1}, \dots, y_{C+M}\}$ from extensive text corpora, where $\mathcal{Y}^- \cap \mathcal{Y} = \emptyset$. Under the assumption that ID samples are more similar to ID labels and less similar to negative labels than OOD samples, NegLabel employs a scoring function to improve OOD detection:
\begin{equation} \label{equ:neglabel_score}
    S_{NegLabel}(\vv) =  \frac{\sum_{i=1}^C {e^{\cos(\vv, \cc_i)/\tau}}}{\sum_{i=1}^C {e^{\cos(\vv, \cc_i)/\tau}} + \sum_{j=C+1}^{C+M} {e^{\cos(\vv, \cc_j)/\tau}}}.
\end{equation}
NegLabel performs excellently on various OOD detection benchmarks by utilizing both ID and  negative text knowledge.

\begin{figure}[tb]
    \centering
    \includegraphics[width=1\columnwidth]{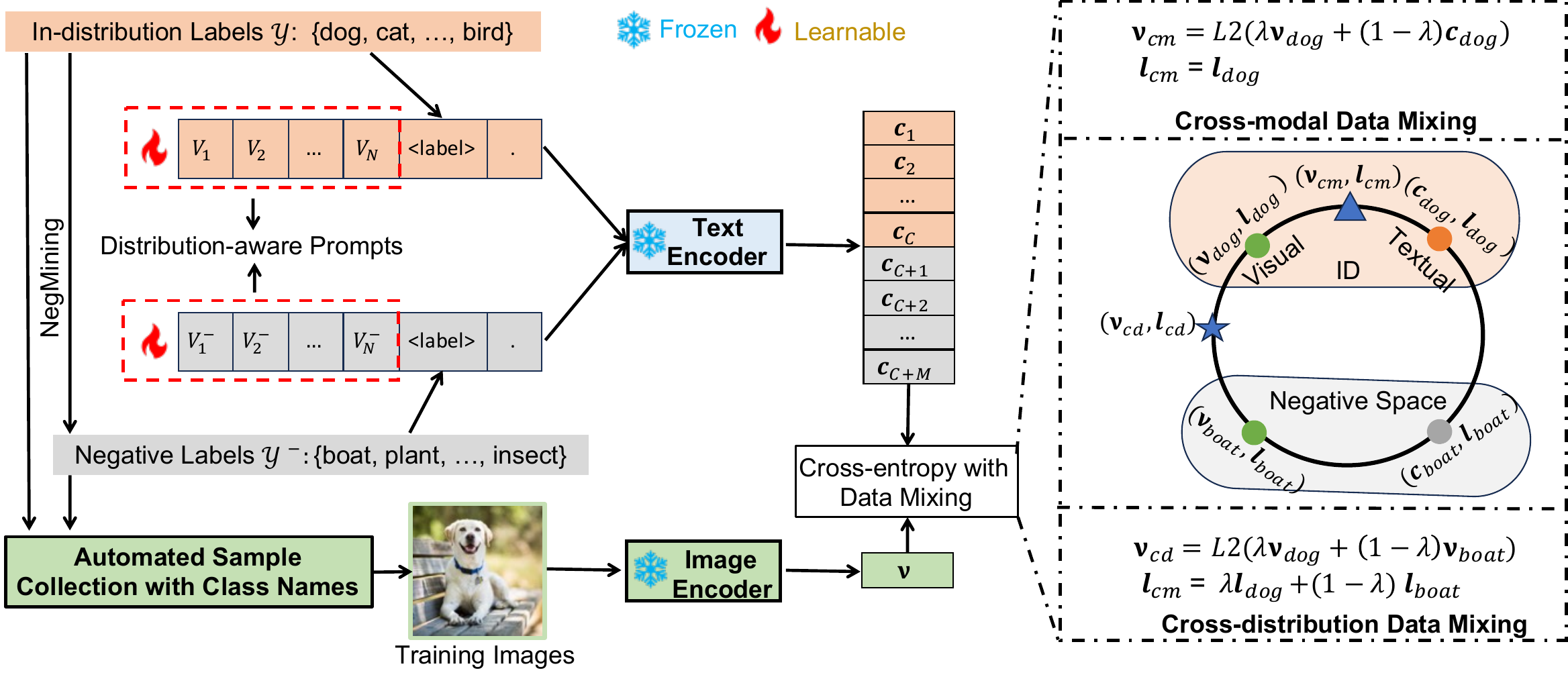}
    \caption{The overall framework of our LAPT method, where $\vv_{dog}/\vv_{boat}$, $\cc_{dog}/\cc_{boat}$, and $\ll_{dog}/\ll_{boat}$ are image features, textual features, and soft labels of dog/boat samples.}
    \label{fig:framework_lapt}
\end{figure}

\subsection{LAPT: Label-driven Automated Prompt Tuning}

Though NegLabel has achieved a notable success, its performance is highly sensitive to the configuration of text prompts. As illustrated in \cref{fig:subtle_text_change}, given an ID dataset like ImageNet, varying text prompts leads to considerable differences in OOD detection outcomes.
For instance, prompts such as `The nice <label>' generally improve performance on far-OOD datasets, while `a photo of a <label>' shows enhanced effectiveness for near-OOD scenarios. 
The choice of prompt can lead to over 10\% fluctuation in the FPR95 metric. Hence, manually setting text prompts for optimal results is a labor-intensive and time-consuming process.

We aim to streamline OOD detection by automating prompt generation with minimal manual intervention. Our method learns continuous prompts using auto-collected samples, needing only ID class names. The automated process is very useful in practical applications. Details of LAPT are provided as follows.

\subsubsection{Distribution-aware prompts.}
Given the ID class names, we first follow NegLabel \cite{jiang2024negative} to mine the negative labels $\mathcal{Y}^-$ from extensive corpus databases. Then, instead of hand-crafting the optimal text prompts, we design a learnable text prompt function in the following form:
\begin{equation} \label{equ:lapt_distribution_aware_prompt}
    \rho_l(\mathrm{LABEL}) = [\V]_1 [\V]_2 \cdots [\V]_N [\mathrm{LABEL}],
\end{equation}
where $[\V]_n \in \mathcal{R}^D (n \in \{1, 2 \cdots, N\})$ represents the learnable context tokens, and $N$ is the total number of context tokens. 

We assess three prompt construction variations for OOD detection: unified, class-specific, and distribution-aware prompts. The unified prompt employs identical context tokens across all classes, and the class-specific prompt utilizes unique tokens for each class, which are explored in the ID classification \cite{zhou2022learning}. Here, we propose a distribution-aware prompt strategy, which differentiates tokens for ID and OOD classes, as shown in Fig. \ref{fig:framework_lapt}.
Our empirical results suggest that the distribution-aware prompt notably enhances the distinction between ID and OOD distributions, as analyzed in Sec. \ref{subsec:analyses_discussion}.

\begin{figure}[tb]
    \centering
    \includegraphics[width=0.83\columnwidth]{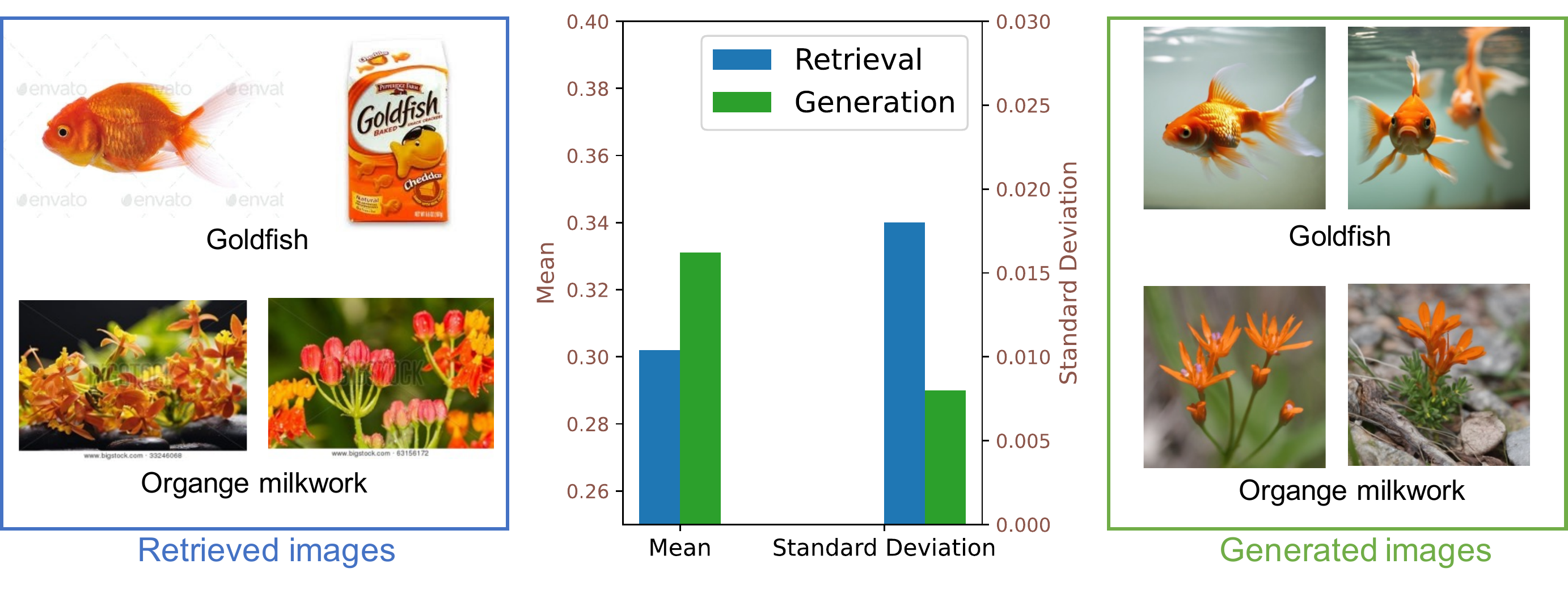}
    \caption{Visualization of retrieved images (left), generated images (right), and the statistic of their cosine similarity to the class labels (middle). A larger mean indicates greater image-label consistency, and a larger standard deviation implies more diversity.}
    \label{fig:col_vis_statistic}
\end{figure}

\subsubsection{Automated sample collection with class labels.}
Traditional prompt learning methods \cite{zhou2022learning,miyai2024locoop} often optimize the prompts with manually collected image samples, which are typically selected from well-constructed benchmark datasets. Such a manual sampling approach is inefficient for negative label-based OOD detection, primarily because mined negative classes are often not included in existing datasets. Moreover, manually collecting and annotating these data is time-consuming and labor-intensive. To mitigate the manual effort and facilitate practical application, we introduce automated sample collection strategies that require only class names, as detailed subsequently.

We introduce two automated sample collection methods for enhancing OOD detection: text-to-image generation and text-based image retrieval. In the generation approach, class names are fed into pre-trained text-to-image models \cite{rombach2022high,podell2023sdxl} to generate synthetic images with varied appearances using different seeds:
\begin{equation}
    \mathcal{X}_{col}^s = \mathrm{T2I}(\mathcal{Y} \cup \mathcal{Y}^-),
\end{equation}
where $\mathcal{X}_{col}^s$ is the collected set of synthetic images and $\mathrm{T2I}(\cdot)$ is the pre-trained generation model.
The performance of OOD detection is closely tied to the fidelity of these generated images, depending on the generative power of the underlying model. By harnessing advanced generative models \cite{rombach2022high,podell2023sdxl}, we are able to produce images that are not only visually convincing but also exhibit strong consistency with the input text. Our empirical findings indicate that these synthetic images can significantly enhance the OOD detection capabilities.

In addition to generating synthetic images with generation models, we also automate the collection of real images through text-based image retrieval. Specifically, we utilize a large-scale real image dataset (\eg, Laion \cite{schuhmann2021laion,schuhmann2022laion}) in conjunction with pre-trained VLMs, such as CLIP, which align images and text within the same feature space. By conducting a search within this space, we retrieve images that are closest to the class names:
\begin{equation}
    \mathcal{X}_{col}^r = \textrm{Clip-retrieval}(\mathcal{Y} \cup \mathcal{Y}^-, \mathrm{WebData}),
\end{equation}
where $\mathcal{X}_{col}^r$ is the retrieved set of real images, $\textrm{Clip-retrieval}$ is the clip-retrieval tool \cite{beaumont-2022-clip-retrieval}, which enables efficient image retrieval from millions of samples, and $\mathrm{WebData}$ is the web-scale real image set.


We capitalize on the strengths of both synthetic image generation and real image retrieval. Synthetic images ensure high textual fidelity and control but may lack variety, while real images offer diversity but can vary in text consistency, as shown in Fig. \ref{fig:col_vis_statistic}. Our strategy prioritizes real images with high text-image alignment, quantified by the cosine similarity between image representation and corresponding text label. If real images fall short in text correlation, indicated by cosine similarity below a threshold, synthetic images are used. This hybrid approach is formalized as follows:
\begin{equation} \label{equ:collection_merge}
    \mathcal{X}_{col} = \{ \x: \x \in \mathcal{X}_{col}^r \text{ if } \cos\left(f_{img}(\x),f_{txt}(y) \right) > \kappa \text{ else } \x \in \mathcal{X}_{col}^s \},
\end{equation}
where $\mathcal{X}_{col}$ represents the final collection of images. The decision to use a real image or a synthetic one is based on whether the cosine similarity exceeds the threshold $\kappa$, which is set as $0.3$ by default following \cite{schuhmann2022laion}.
This hybrid approach optimizes the balance between image diversity and text-image consistency, enhancing the robustness of our method.




\subsubsection{Prompt tuning with cross-modal and cross-distribution mixing.} 
Given the training images collected on a per-class basis, the most straightforward learning approach is to use a classification objective, which is commonly achieved with the following cross-entropy loss function:
\begin{equation} \label{equ:vanilla_ce}
    \mathcal{L} = - \frac{1}{|\mathcal{X}_{col}|} \sum_{\x \in \mathcal{X}_{col}} \sum_{i=1}^{C+M} \ll_i \log \frac{e^{\cos\left(\vv, (f_{txt}(\rho_l(y))\right)/\tau}}{\sum_{i=1}^{C+M} {e^{\cos(\vv, \cos(f_{txt}(\rho_l(y)))/\tau}}}, \vv = f_{img}(\x)
\end{equation}
where $y$ is the class name of $\x$.
$\ll_i$ is the $i$-th entry of $\ll \in \mathcal{R}^{C+M}$, the softlabel of sample $\x$. 
In Sec. \ref{subsec:analyses_discussion}, we will demonstrate that this training objective aligns effectively with our proposed distribution-aware prompt construction and automated sample collection strategies. 

\textbf{\textit{Cross-modal mixing.}} While the simple classification loss has achieved commendable results, we could further enhance it by addressing potential sample noise and exploring broader data space through carefully designed data mixing strategies. Despite the efforts to collect clean images that match their respective class names, noise is inevitable in the collected images. To mitigate the impact of image noise, we introduce a cross-modal mixing strategy to neutralize potential noise in the images. Specifically, for a given training image $\x$ of class $y$, we introduce the refined training feature $\vv_{cm}$ as follows:
\begin{align} \label{equ:cross_modal_mix}
    \vv_{cm} &= \mathrm{L_2}\left(\lambda_{\alpha} f_{img}(\x) + (1-\lambda_{\alpha}) f_{txt}\left(\rho(y)\right)\right), \\
    \ll_{cm} &= \ll,
\end{align}
where $\lambda_{\alpha} \sim \mathrm{Beta}(\alpha, \alpha) \in [0,1]$ with $\alpha \in (0,\infty)$. $\mathrm{L_2}(\cdot)$ represents the $L_2$ normalization along feature dimension and $\rho(\cdot)$ is the hand-crafted text prompts \cite{radford2021learning}. 
The softlabel is unchanged after mixing since the image $\x$ and its label $y$ belong to the same class. 
When $\lambda_{\alpha}=1$, we adopt the image features as the training examples, and we use the text features as the training examples when $\lambda_{\alpha}=0$. 
We denote the cross-entropy loss of \cref{equ:vanilla_ce} with $\vv_{cm}$ and $\ll_{cm}$ as $\mathcal{L}_{cm}$.
By mixing textual and visual representations, we aim to create a more robust feature set that can help the model learn to focus on relevant features while disregarding noisy information. This strategy not only reinforces the model's resistance to image noise but also encourages it to leverage complementary information from both modalities.

\textbf{\textit{Cross-distribution mixing.}} In addition to using cross-modal data mixing to mitigate image noise, we employ the data mixing strategy to explore a broader feature space. In the vanilla NegLabel approach \cite{jiang2024negative}, the selected negative classes are typically far from the ID distribution, leaving a substantial intermediate area between ID and negative regions underutilized. To enhance the utilization of these spaces, we introduce a cross-distribution mixing strategy, which combines features and corresponding labels of ID and negative samples to create new training samples.
This mixing strategy effectively bridges the distribution gap, generating a continuous spectrum of features that extend from ID to negative areas.
The mixing process can be formalized as follows:
\begin{align} \label{equ:cross_dis_mix}
    \vv_{cd} &= \mathrm{L2}\left(\lambda_{\beta} f_{img}(\x_{id}) + (1-\lambda_{\beta})  f_{img}(\x_{ood})\right), \\
    \ll_{cd} &= \lambda_{\beta} \ll_{id} + (1-\lambda_{\beta})  \ll_{ood},
\end{align}
where $\x_{id}$ and $\x_{ood}$ are collected positive and negative images, and $\ll_{id}$ and $\ll_{ood}$ are soft labels of $\x_{id}$ and $\x_{ood}$, respectively.
$\lambda_{\beta} \sim \mathrm{Beta}(\beta, \beta) \in [0,1]$ with $\beta \in (0,\infty)$.
This method not only allows the model to become more aware of the intermediate space between ID and negative regions, but also encourages it to learn more discriminative features that can better generalize to new and unseen OOD samples.
We denote the cross-entropy loss of \cref{equ:vanilla_ce} with $\vv_{cd}$ and $\ll_{cd}$ as $\mathcal{L}_{cd}$. The final objective is constructed as:
\begin{equation} \label{equ:lapt_all}
    \mathcal{L}_{all} = \mathcal{L} + \mathcal{L}_{cm} + \mathcal{L}_{cd}.
\end{equation}

\textbf{\textit{Remarks.}} Our proposed data mixing strategies are distinct from existing ones \cite{zhang2017mixup,verma2019manifold,yang2023mixood,Zhang_2023_WACV} in both implementation and objectives. Traditional data mixing strategies typically aim to enhance ID classification performance by mixing randomly selected ID images \cite{zhang2017mixup,verma2019manifold}. In contrast, our approach is carefully designed to include multi-modal and multi-distribution data in the mixing process to enhance the OOD detection capability.
While there are some methods that utilize data mixing for OOD detection, they usually mix ID data to simulate OOD representations \cite{yang2023mixood} or combining manually collected ID and OOD data \cite{Zhang_2023_WACV}. These methods, however, do not exploit the textual knowledge and require manually labeled samples. 
Our cross-modal mixing strategy lowers image noise by blending samples from the same class but across different modalities, creating more robust features. Meanwhile, our cross-distribution mixing scheme goes a step further by mixing automatically collected positive and negative data, allowing us to explore the space that lies between ID and negative regions. This exploration enables our model to better understand and characterize the boundary between ID and OOD, leading to more effective detection of anomalies.
Our methods are not just mixing data; they are deliberately designed to align with our label-driven automated prompt learning objective. By addressing noise and exploring intermediate feature space through mixing, we develop a tailored solution that boosts the automated prompt tuning for OOD detection.


In the test stage, we adopt the OOD score of \cref{equ:neglabel_score} following \cite{jiang2024negative}. The overall pipeline of our automated prompt learning is summarized in \cref{algorithm:lapt}.

\begin{algorithm} [tb]
\caption{Label-driven Automated Prompt Tuning (LAPT)} \label{algorithm:lapt}
\begin{algorithmic}[1] 
    \REQUIRE ID label space $\mathcal{Y}$ 
    \STATE $\mathcal{Y}^- \leftarrow$ NegMine($\mathcal{Y}$)
    \STATE Constructing distribution-aware prompts $\rho_l$ with labels $\mathcal{Y} \cup \mathcal{Y}^-$ using Eq. (\ref{equ:lapt_distribution_aware_prompt})
    \STATE $\mathcal{X}_{col} \leftarrow$ Automated sample collection with labels $\mathcal{Y} \cup \mathcal{Y}^-$ using Eq. (\ref{equ:collection_merge}) 
    \STATE Prompt tuning with cross-modal and cross-distribution data mixing using Eq. (\ref{equ:lapt_all})
    \RETURN Learned prompts $\rho_l$
\end{algorithmic}
\end{algorithm}

\section{Experiments}

\subsection{Setup}
\textbf{Datasets and benchmarks.}
Our validation primarily uses the ImageNet-1k dataset \cite{deng2009imagenet} as ID data, and more evaluations on smaller-scaled ID datasets are provided in the Supplementary Materials. Aligning with standards from prior research \cite{huang2021mos,ming2022delving,jiang2024negative}, we use four diverse datasets of iNaturalist \cite{van2018inaturalist}, SUN \cite{xiao2010sun}, Places \cite{zhou2017places}, and Textures \cite{cimpoi2014describing} as OOD test sets.
Additionally, we assess our method on the OpenOOD benchmark \cite{zhang2023openood,yang2022openood}, which differentiates OOD datasets into near-OOD (\eg, SSB-hard \cite{vaze2021open}, NINCO \cite{bitterwolf2023ninco}) and far-OOD (\eg, iNaturalist \cite{van2018inaturalist}, Textures \cite{cimpoi2014describing}, OpenImage-O \cite{wang2022vim}) based on semantic similarity or difficulty. This grouping allows for a comprehensive evaluation of OOD detectors against various OOD sample types.

In addition to conventional OOD detection that solely considers semantic shifts, we consider a full-spectrum OOD detection problem \cite{yang2023full}, which additionally accounts for non-semantic covariate shift generalization. An optimal system should not only be capable of identifying semantically shifted OOD samples but should also exhibit robustness to non-semantic covariate shifted OOD samples. Following the guidelines in \cite{zhang2023openood}, we adopt ImageNet-C \cite{hendrycks2019benchmarking}, ImageNet-R \cite{hendrycks2021many}, and ImageNet-V2 \cite{recht2019imagenet} as the covariate-shifted ID data in the full-spectrum OOD detection evaluation.

Our evaluation metrics include (1) FPR95, the false positive rate of OOD samples with ID detection rate at 95\%; (2) AUROC, the area under the receiver operating characteristic curve; and (3) ID classification accuracy (ID ACC).



\textbf{Implementation details.}
We adopt the visual encoder of VITB/16 pretrained by CLIP \cite{radford2021learning}. 
For automated sample collection, we instantiate the text-to-image generation model with SDXL \cite{podell2023sdxl} and employ a subset of the Laion400M dataset \cite{schuhmann2022laion},  approximately 60 million samples, as our WebData.
Our model training is conducted using the SGD optimizer with a learning rate schedule that follows cosine annealing, starting from an initial learning rate of 1e-2. We set the batch size to $32$ and train the model for $10$ epochs in all experiments.



\begin{table}[tb] \scriptsize
\centering
\caption{OOD detection results with ID data of ImageNet-1k and four OOD datasets by using the VITB/16 CLIP encoder.} \label{tab:four_ood_datasets}
\begin{tabular}{lcccccccc|cc}
\toprule
\multicolumn{11}{c}{OOD datasets}  \\
\multicolumn{1}{c}{\multirow{2}{*}{Methods}} & \multicolumn{2}{c}{INaturalist} & \multicolumn{2}{c}{Sun} & \multicolumn{2}{c}{Places} & \multicolumn{2}{c}{Textures} & \multicolumn{2}{c}{Average} \\ \cline{2-3} \cline{4-5} \cline{6-7} \cline{8-9} \cline{10-11}
 & \tiny AUROC$\uparrow$ & \tiny FPR95$\downarrow$& \tiny AUROC$\uparrow$ & \tiny FPR95$\downarrow$& \tiny AUROC$\uparrow$ & \tiny FPR95$\downarrow$& \tiny AUROC$\uparrow$ & \tiny FPR95$\downarrow$ & \tiny AUROC$\uparrow$ & \tiny FPR95$\downarrow$        \\
 \midrule
 \multicolumn{11}{c}{\textbf{Requires Manually Labeled Training Samples}} \\
MSP \cite{hendrycks2016baseline}    &    87.44 & 58.36 & 79.73 & 73.72 & 79.67 & 74.41 & 79.69 & 71.93 & 81.63 & 69.61   \\
ODIN \cite{liang2017enhancing}   &    94.65 & 30.22 & 87.17 & 54.04 & 85.54 & 55.06 & 87.85 & 51.67 & 88.80 & 47.75   \\
Energy \cite{liu2020energy}  &    95.33 & 26.12 & 92.66 & 35.97 & 91.41 & 39.87 & 86.76 & 57.61 & 91.54 & 39.89 \\
GradNorm \cite{huang2021importance} &   72.56 & 81.50 & 72.86 & 82.00 & 73.70 & 80.41 & 70.26 & 79.36 & 72.35 & 80.82 \\
ViM \cite{wang2022vim}   &    93.16 & 32.19 & 87.19 & 54.01 & 83.75 & 60.67 & 87.18 & 53.94 & 87.82 & 50.20 \\
KNN \cite{sun2022out}   &    94.52 & 29.17 & 92.67 & 35.62 & 91.02 & 39.61 & 85.67 & 64.35 & 90.97 & 42.19 \\
VOS \cite{du2022unknown}   &    94.62 & 28.99 & 92.57 & 36.88 & 91.23 & 38.39 & 86.33 & 61.02 & 91.19 & 41.32 \\
NPOS \cite{tao2023non}  &    96.19 & 16.58 & 90.44 & 43.77 & 89.44 & 45.27 & 88.80 & 46.12 & 91.22 & 37.93\\
LSN \cite{nie2023out} & 95.83 & 21.56 & 94.35 & 26.32 & 91.25 & 34.48 & 90.42 & 38.54 & 92.96 & 30.22 \\
LoCoOp \cite{miyai2024locoop} & 96.86 & 16.05 & 95.07 & 23.44 & 91.98 & 32.87 & 90.19 & 42.28 & 93.52 & 28.66 \\
 \midrule
  \multicolumn{11}{c}{\textbf{Does Not Require Manually Labeled Training Samples}} \\
Mahalanobis \cite{lee2018simple}      & 55.89 & 99.33 & 59.94 & 99.41 & 65.96 & 98.54 & 64.23 & 98.46 & 61.50 & 98.94 \\
Energy \cite{liu2020energy}           & 85.09 & 81.08 & 84.24 & 79.02 & 83.38 & 75.08 & 65.56 & 93.65 & 79.57 & 82.21 \\
ZOC \cite{esmaeilpour2022zero}               & 86.09 & 87.30 & 81.20 & 81.51 & 83.39 & 73.06 & 76.46 & 98.90 & 81.79 & 85.19 \\
MCM \cite{ming2022delving} & 94.59 & 32.20 & 92.25 & 38.80 & 90.31 & 46.20 & 86.12 & 58.50 & 90.82 & 43.93 \\
CLIPN \cite{wang2023clipn}            & 95.27 & 23.94 & 93.93 & 26.17 & \textbf{92.28} & 33.45 & 90.93 & 40.83 & 93.10 & 31.10 \\
NegLabel \cite{jiang2024negative} &99.49&1.91&95.49& 20.53 & 91.64 & 35.59 & 90.22 & 43.56 & 94.21 & 25.40 \\
\rowcolor{HighLight} \textbf{LAPT (Ours)} & \textbf{99.63} & \textbf{1.16} & \textbf{96.01} & \textbf{19.12} & 92.01 & \textbf{33.01} & \textbf{91.06} & \textbf{40.32} & \textbf{94.68} & \textbf{23.40} \\
\bottomrule
\end{tabular}
\end{table}

\subsection{Main Results}

\textbf{Results on four OOD datasets.}
As illustrated in Tab. \ref{tab:four_ood_datasets}, our approach consistently surpasses existing methods. Specifically, we re-implemented traditional methods \cite{hendrycks2016baseline,liang2017enhancing,liu2020energy,huang2021importance,wang2022vim,sun2022out,du2022unknown,tao2023non} by fine-tuning CLIP-encoders with labeled training samples from ImageNet in accordance with \cite{tao2023non}. Without utilizing any manually annotated data, our method significantly outperforms these methods, corroborating the effectiveness of our proposed VLMs adaptation technique. Notably, our approach exceeds the performance of NegLabel \cite{jiang2024negative}, confirming the superiority of learnable prompts over hand-crafted prompts. The OOD samples in these datasets have a relatively large semantic difference from the ID data (\eg, textures vs. animals), making the distinction relatively easy. On more challenging datasets, where the OOD data are semantically closer to the ID data, our advantages are even more pronounced, as shown in Tabs. \ref{tab:openood_lapt} and \ref{tab:openood_lapt_fsood}.


\textbf{Results on OpenOOD benchmark.} 
A more comprehensive evaluation of our method is provided in Tab. \ref{tab:openood_lapt} on the OpenOOD benchmark \cite{yang2022openood,zhang2023openood}. 
Compared to its closest competitor NegLabel, our method not only enhances the OOD detection performance but also improves the ID classification accuracy. In terms of OOD detection, our approach achieves a slight improvement in the relatively easy far-OOD scenarios (\eg, 0.96\% AUROC) and a substantial improvement in the more challenging near-OOD settings (\eg, 5.71\% AUROC), highlighting the outstanding effectiveness of our method in challenging contexts.
The competitors with manually labeled training samples are quoted from \cite{yang2022openood,zhang2023openood}, where models are trained with full ImageNet training data.

\begin{table}[tb]
    \centering
\caption{OOD detection results on the OpenOOD benchmark, where ImageNet-1k is adopted as ID dataset. Full results are available in the Supplementary Materials.}\label{tab:openood_lapt}
\begin{tabular}{l|cc|cc|c}
\toprule
\multirow{2}{*}{ Methods } & \multicolumn{2}{|c|}{ FPR95 $\downarrow$} & \multicolumn{2}{|c|}{$\mathrm{AUROC} \uparrow$} & ACC $\uparrow$ \\
\cline{2-5}
& Near-OOD & Far-OOD & Near-OOD & Far-OOD & ID \\
\midrule
 \multicolumn{6}{c}{\textbf{Requires Manually Labeled Training Samples}} \\
GEN \cite{liu2023gen}  & -- & -- & 78.97 & 90.98 & 81.59 \\
AugMix \cite{hendrycks2019augmix} + ReAct \cite{sun2021react} & -- & -- & 79.94 & 93.70 & 77.63 \\
RMDS \cite{ren2021simple} & -- & -- & 80.09 & 92.60 & 81.14  \\
AugMix \cite{hendrycks2019augmix} + ASH \cite{djurisic2022extremely}   & -- & -- & 82.16 & 96.05 & 77.63 \\
\midrule
\multicolumn{6}{c}{\textbf{Does Not Require Manually Labeled Training Samples}} \\
MCM \cite{ming2022delving}        & 79.02 & 68.54 & 60.11 & 84.77 & 66.28 \\
NegLabel \cite{jiang2024negative} & 68.18 & 27.34 & 76.92 & 93.30 & 66.82\\
\rowcolor{HighLight} \textbf{LAPT (Ours)} & \textbf{58.94} & \textbf{24.86} & \textbf{82.63} & \textbf{94.26} & \textbf{67.86} \\
\bottomrule
\end{tabular}
\end{table}

\textbf{Full-spectrum results on OpenOOD benchmark.} 
In addition to the capabilities of semantic OOD detection and ID sample recognition, we also analyze the generalization ability of our method under covariate shift \cite{zhou2023context,zhou2023self,zhang2020unsupervised,zhang2020label} on the full-spectrum OOD detection task \cite{yang2023full} using the OpenOOD benchmark \cite{yang2022openood,zhang2023openood}. As illustrated in Tab. \ref{tab:openood_lapt_fsood}, our method enhances all the three capacities, achieving improved results across the board. Similar to the findings in Tab. \ref{tab:openood_lapt}, the improvement is more pronounced in the challenging near-OOD scenarios.


\begin{table}[tb]
    \centering
\caption{Full-specturm OOD detection results on the OpenOOD benchmark, where ImageNet-1k, ImageNet-C, ImageNet-R, ImageNet-V2 are adopted as ID datasets. }\label{tab:openood_lapt_fsood}
\begin{tabular}{l|cc|cc|c}
\toprule
\multirow{2}{*}{ Methods } & \multicolumn{2}{|c|}{ FPR95 $\downarrow$} & \multicolumn{2}{|c|}{$\mathrm{AUROC} \uparrow$} & ACC $\uparrow$ \\
\cline{2-5}
& Near-OOD & Far-OOD & Near-OOD & Far-OOD & ID \\
\midrule
\multicolumn{6}{c}{\textbf{Requires Manually Labeled Training Samples}} \\
DeepAug \cite{hendrycks2021many} + SHE \cite{zhang2022out} & 83.26 & 68.70 & 68.27 & 78.85 &  57.82 \\
StyAug \cite{geirhos2018imagenet} + GradNorm \cite{huang2021importance} & 87.14 & 58.82 & 65.27 & 81.62 & 55.44 \\
AugMix \cite{hendrycks2019augmix} + SHE \cite{zhang2022out} & 84.45 & 60.26 & 69.66 & 83.06 &  57.46 \\
ASH \cite{djurisic2022extremely} & 93.27 & 59.56 & 60.52 & 86.75 &  54.35 \\
LSA \cite{lu2023likelihood} & 70.56 & 48.06 & 78.22 & 86.85 & 61.74 \\
ISH + SCALE \cite{xu2023scaling} & -- & -- &  68.04	& 89.46	& 54.88 \\
\midrule
\multicolumn{6}{c}{\textbf{Does Not Require Manually Labeled Training Samples}} \\
MCM \cite{ming2022delving}        & 85.37 & 69.87 & 58.97 & 77.11 & 57.69 \\
NegLabel \cite{jiang2024negative} & 76.25 & 33.30 & 72.77 & 92.02  & 60.11  \\
\rowcolor{HighLight} \textbf{LAPT (Ours)} & \textbf{71.18} & \textbf{33.07} & \textbf{74.77} & \textbf{92.14} & \textbf{61.17} \\
\bottomrule
\end{tabular}
\end{table}

\subsection{Analyses and Discussions} \label{subsec:analyses_discussion}
The analyses are conducted on the OpenOOD dataset with a VITB/16 encoder.

\begin{figure}[tb]
	\centering
	\begin{subfigure}{0.242\linewidth}
		\includegraphics[width=\linewidth]{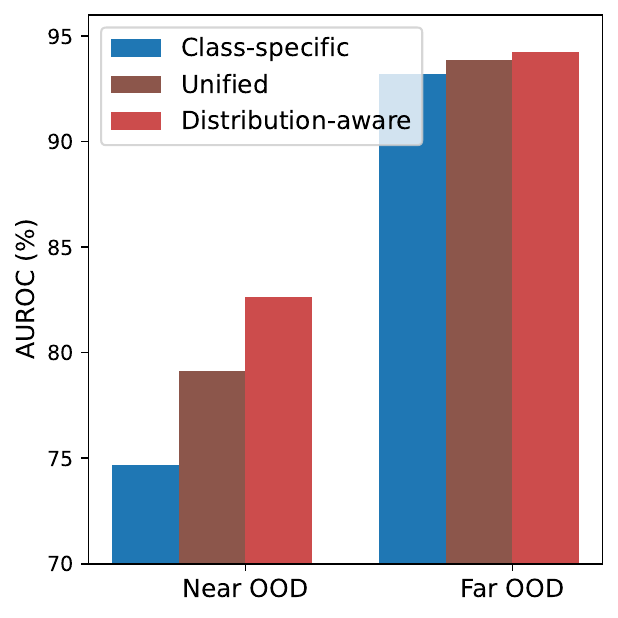}
		\caption{Types}
		\label{Fig:prompt_types}
	\end{subfigure}
	\hfill
	\begin{subfigure}{0.242\linewidth}
		\includegraphics[width=\linewidth]{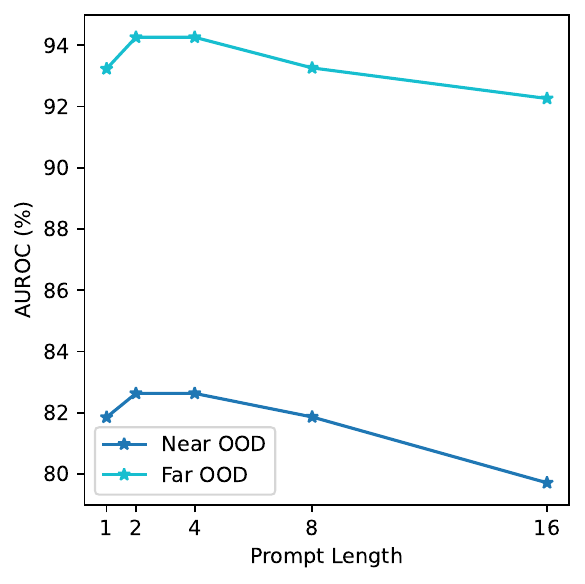}
		\caption{Length $N$}
		\label{Fig:prompt_length}
	\end{subfigure}
	\hfill
	\begin{subfigure}{0.242\linewidth}
		\includegraphics[width=\linewidth]{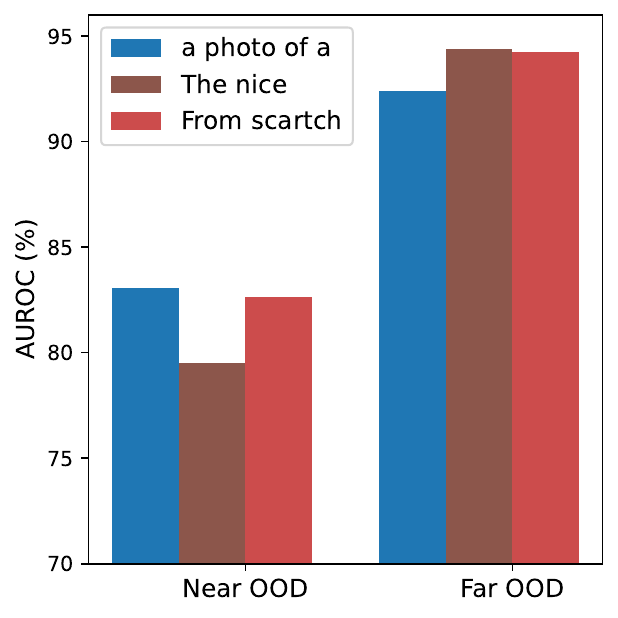}
		\caption{Initializations}
		\label{Fig:prompt_init}
	\end{subfigure}
	\hfill
	\begin{subfigure}{0.242\linewidth}
		\includegraphics[width=\linewidth]{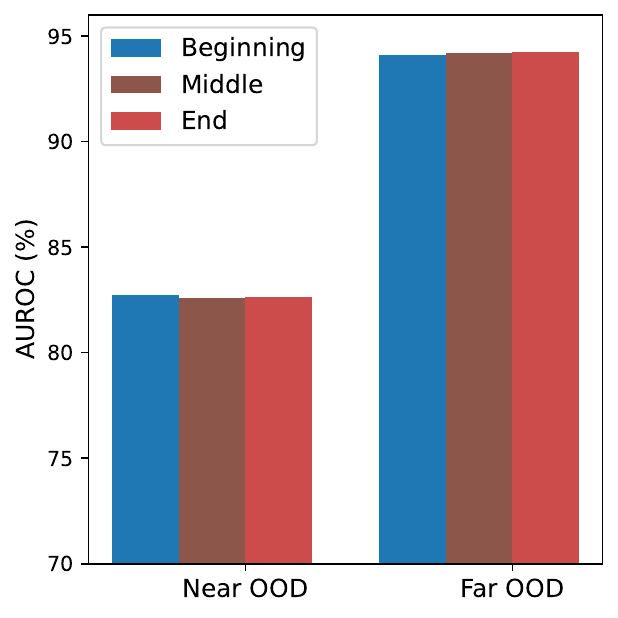}
		\caption{Positions}
		\label{Fig:prompt_position}
	\end{subfigure}
	\caption{Analyses on prompt (a) types, (b)  length, (b) initializations, and (d) positions.}
	\label{Fig:prompt_analyses}
\end{figure}

\textbf{Prompt construction.}
We conduct comprehensive analyses on prompt construction as demonstrated in \cref{Fig:prompt_analyses}. As shown in \cref{Fig:prompt_types}, our proposed distribution-aware prompts are well-matched for OOD detection, outperforming both class-specific and unified prompts. Excessively long prompts slightly impair experimental outcomes, as evidenced in  \cref{Fig:prompt_length}, possibly due to the increased capacity overfitting noise presented in the training images. Consequently, we set the prompt length as $N=2$ by default. As indicated in  \cref{Fig:prompt_init}, initializing prompts with manually crafted text introduces textual bias. For instance, `The nice' yields better results for far-OOD but proves suboptimal for near-OOD, a bias that is perpetuated in our method when learnable prompts are initialized with `The nice'. We discover that a random initialization, which does not introduce any prior, performs consistently well across different OOD tasks and is thus adopted as the default practice. Finally, in \cref{Fig:prompt_position}, we observe that the placement of label tokens has a minimal impact on the results; hence, we by default place the class label at the end of the prompt, as shown in Eq. (\ref{equ:lapt_distribution_aware_prompt})

\begin{figure}[tb]
	\centering
	\begin{subfigure}{0.242\linewidth}
		\includegraphics[width=\linewidth]{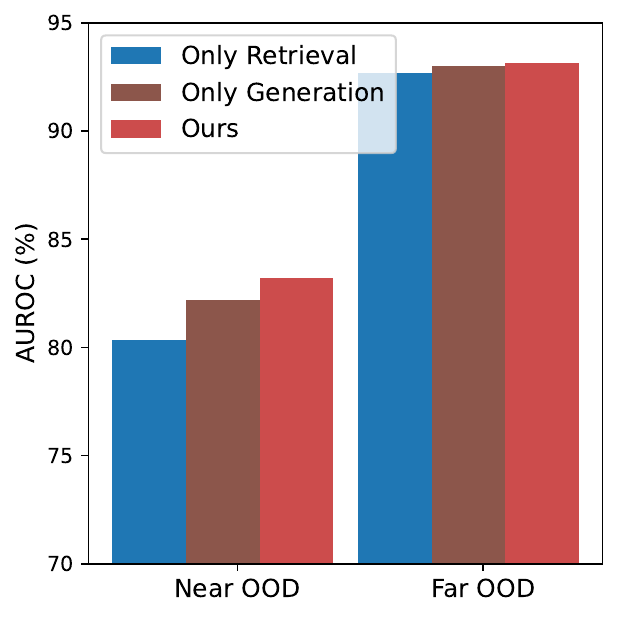}
		\caption{Collection Methods}
		\label{Fig:collection_methods}
	\end{subfigure}
	\hfill
	\begin{subfigure}{0.242\linewidth}
		\includegraphics[width=\linewidth]{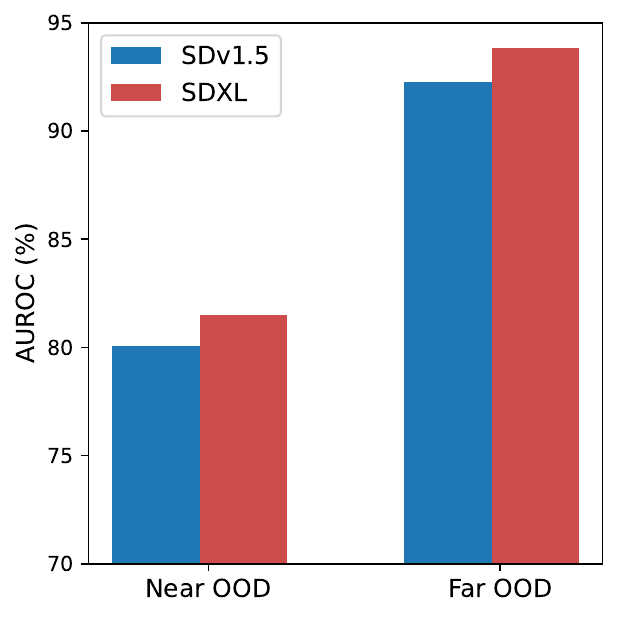}
		\caption{Generators}
		\label{Fig:generators_type}
	\end{subfigure}
	\hfill
	\begin{subfigure}{0.242\linewidth}
		\includegraphics[width=\linewidth]{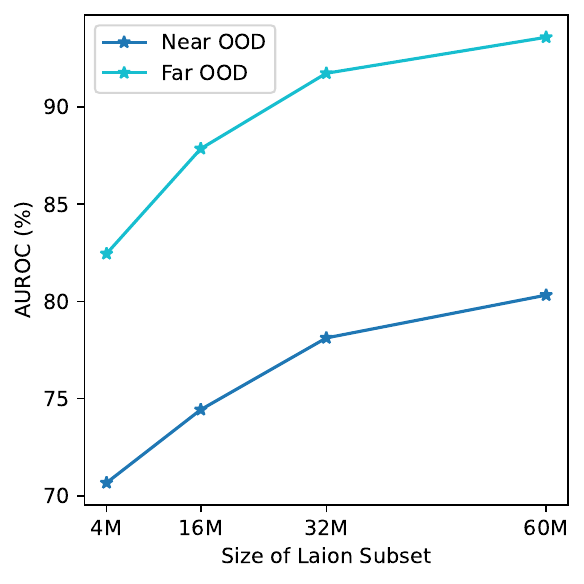}
		\caption{WebData Scale}
		\label{Fig:laion_size}
	\end{subfigure}
	\hfill
	\begin{subfigure}{0.242\linewidth}
		\includegraphics[width=\linewidth]{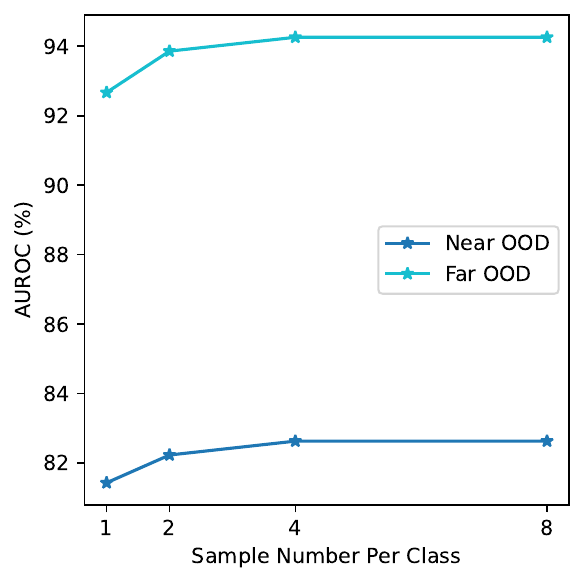}
		\caption{Sample Number}
		\label{Fig:sample_number}
	\end{subfigure}
	\caption{Analyses on (a) image collection methods, (b) text-to-image generators, (c) data scale of the WebData space, and (d) number of training samples per class.}
	\label{Fig:collection_strategies}
\end{figure}

\textbf{Automated sample collection.}
As shown in Fig. \ref{Fig:collection_methods}, the synthetic image generation strategy outperforms realistic image retrieval, possibly because the generator has been exposed to a vast amount of training data (\eg, 5 billion images), which greatly surpasses the scale of our utilized WebData (\eg, 60 million images). The best results are obtained by synergistically leveraging both synthetic and real data, confirming their complementary nature.
In terms of image generation, employing a stronger image generator can result in better performance, as shown in Fig. \ref{Fig:generators_type}. Regarding image retrieval, utilizing a larger-scale WebData can yield better outcomes, as presented in Fig. \ref{Fig:laion_size}. Finally, we find in Fig. \ref{Fig:sample_number} that the model's performance does not continue to increase with the addition of more training samples.
This may be because it is difficult to collect a high-quality training set that is semantically consistent with the text and diverse. With the relatively small size of WebData, it is challenging to ensure semantic consistency, and while the semantic consistency of generated images may be good, their diversity is limited, as analyzed in Fig. \ref{fig:col_vis_statistic}.
How to automatically collect high-quality images for effective model training remains an enduring and challenging problem.


More analyses on stage ablation, the impact of text prompts in data collection, data mixing strategies, hyper-parameters in data mixing, different VLMs, visualization of learned prompts, and time complexity are provided in the Supplementary Materials.







\section{Conclusion and Limitations}
We developed a novel label-driven automated prompt tuning (LAPT) approach that autonomously configures effective prompts for OOD detection. Requiring only ID class names, our method can automatically mine OOD class names, collect training images per class, and learn effective prompts, thereby outperforming manually crafted prompts with minimal manual intervention. The automatically learned prompts not only enhanced the distinction between ID and OOD samples but also improved the ID classification accuracy and the strengthened generalization robustness to covariate shifts.

The efficacy of our approach was highly dependent on the 
quality of collected training samples.
Employing more advanced generation techniques to create a diverse set of synthetic images \cite{dunlap2023diversify} or expanding the retrieval space \cite{schuhmann2021laion} to obtain real images that match better the class names could yield improved results. These possibilities present exciting avenues for future research.

\bibliographystyle{splncs04}
\bibliography{egbib}
\end{document}